\documentclass{article}

\usepackage{natbib}
\usepackage{microtype}
\usepackage{subfigure}
\usepackage{booktabs} %

\usepackage{hyperref}

\usepackage{url}
\usepackage{amsfonts}
\usepackage{graphicx}
\usepackage{amsmath,amssymb,amsthm}
\usepackage{mathtools}
\usepackage[nameinlink]{cleveref}

\usepackage[margin=1in]{geometry}  %

\newtheorem{theorem}{Theorem}

\newtheorem{corollary}{Corollary}[theorem]

\crefname{theorem}{Theorem}{Theorems}
\crefname{corollary}{Corollary}{corollaries}
\crefname{equation}{Eq.}{Equations}
\crefname{table}{Table.}{Tables}
\crefname{figure}{Fig.}{Figures}
\crefname{section}{Section}{Sections}

\newcommand{\KL}{\mathrm{KL}}

\newcommand{\sign}{\mathrm{sign}}

\newcommand{\bbE}{\mathop{\mathbb{E}}}

\newcommand{\idh}{\mathrm{d}{\mathbf{h}}}

\newcommand{\bbI}{\mathbb I}

\newcommand{\bbN}{\mathbb N}

\newcommand{\bbR}{\mathbb R}

\newcommand{\calC}{\mathcal C}
\newcommand{\calD}{\mathcal D}

\newcommand{\calH}{\mathcal H}

\newcommand{\calN}{\mathcal N}
\newcommand{\calO}{\mathcal O}
\newcommand{\calP}{\mathcal P}
\newcommand{\calQ}{\mathcal Q}

\newcommand{\calS}{\mathcal S}

\newcommand{\calV}{\mathcal V}

\newcommand{\calX}{\mathcal X}
\newcommand{\calY}{\mathcal Y}

\newcommand{\bfh}{\mathbf h}
\newcommand{\bfi}{\mathbf i}

\newcommand{\bfo}{\mathbf o}

\newcommand{\bfx}{\mathbf x}

\newcommand{\bfI}{\mathbf I}

\newcommand{\bfW}{\mathbf W}

\title{PAC-Bayes Analysis of Sentence Representation}

\author{
    Kento Nozawa \\
    The University of Tokyo \& RIKEN \\
    \texttt{\href{mailto:nozawa@ms.k.u-tokyo.ac.jp}{nozawa@ms.k.u-tokyo.ac.jp}} \\
    \texttt{\href{https://nzw0301.github.io}{https://nzw0301.github.io}} \\
    \and
    Issei Sato \\
    The University of Tokyo \& RIKEN \\
    \texttt{\href{mailto:sato@k.u-tokyo.ac.jp}{sato@k.u-tokyo.ac.jp}} \\
}
\begin{document}

\maketitle

\begin{abstract}
    Learning sentence vectors from an unlabeled corpus has attracted attention because such vectors can represent sentences in a lower dimensional and continuous space.
    Simple heuristics using pre-trained word vectors are widely applied to machine learning tasks.
    However, they are not well understood from a theoretical perspective.
    We analyze learning sentence vectors from a transfer learning perspective by using a PAC-Bayes bound that enables us to understand existing heuristics.
    We show that simple heuristics such as averaging and inverse document frequency weighted averaging are derived by our formulation.
    Moreover, we propose novel sentence vector learning algorithms on the basis of our PAC-Bayes analysis.
\end{abstract}

\section{Introduction}
Representation learning~\citep{Bengio2013IEEE} is a class of the most fundamental machine learning tasks.
It aims to obtain re-usable representations that capture useful features from a massive amount of data.
Usually, a simple representation for a data sample is a fixed length $d$-dimensional real-valued vector.
Therefore, it is used as the feature vector of several machine learning tasks, such as classification, clustering, and visualization.
Representation learning by solving supervised tasks requires a lot of labeled data, especially, when using deep neural network models.
For example, DeCAF~\citep{Donahue2014ICML} image feature representations are based on AlexNet~\citep{Krizhevsky2012NeurIPS} trained on ImageNet~\citep{Deng2009CVPR},
and InferSent~\citep{Conneau2017EMNLP} sentence feature representations based on deep neural networks trained on the SNLI dataset~\citep{Bowman2015EMNLP}, which consists of 570k labeled sentence pairs.
\citet{Kornblith2018arXiv} report a strong correlation between a source task's performance and a target task's performance with representations trained on the source task regarding image classification tasks.
Therefore, a large amount of labeled data is one of the most critical factors of supervised representation learning.
In contrast, unsupervised representation learning does not require label information.
In addition, we easily apply these algorithms not only to labeled datasets, e.g., ImageNet and SNLI, but also large unlabeled datasets, e.g., images and text collected by a web crawler.

Unsupervised word representation learning~\citep{Collobert2011JMLR, Minh2013NIPS, Mikolov2013NIPS, Pennington2014EMNLP, Bojanowski2017TACL} is one of the most successful representation learning tasks.
These word representations often improve the performance of other natural language processing tasks,
such as text classification~\citep{Mikolov2018LREC},
neural machine translation~\citep{Qi2018NAACL},
word similarity and analogy task \citep{Bojanowski2017TACL},
and several benchmark tasks~\citep{Peters2018NAACL, Peters2018EMNLP}.
In addition, we can easily use publicly available word vectors trained on unlabeled text datasets,
e.g.,
\texttt{SENNA}\footnote{\url{https://ronan.collobert.com/senna/}}~\citep{Collobert2011JMLR},
\texttt{word2vec}\footnote{\url{https://code.google.com/archive/p/word2vec/}} \citep{Mikolov2013NIPS},
\texttt{GloVe}\footnote{\url{https://nlp.stanford.edu/projects/glove/}}~\citep{Pennington2014EMNLP},
and \texttt{fastText}\footnote{\url{https://fasttext.cc/docs/en/crawl-vectors}}~\citep{LREC2018Grave}.

Unsupervised sentence representation learning~\citep{Le2014ICML, Kiros2015NIPS} is a more complex task than word representation learning because a sentence's meaning is determined by several factors, such as words, the word order, and grammar.
The simplest algorithm is to average pre-trained word vectors over words appearing in a sentence.
Even if we use models based on recurrent neural networks with many learning parameters,
simpler models based on pre-trained word vectors beat these large models~\citep{Hill2016NAACL, Wieting2016ICLR, Arora2017ICLR, Shen2018ACL, Wieting2019ICLR}.
Unlike theoretical perspectives on word representation learning~\citep{Levy2014NIPS, Arora2016TACL, Melamud2017ACL},
sentence representation learning is not understood well theoretically.
As a first step toward theoretical understanding of sentence representation learning,
we analyze simple heuristics from a transfer learning perspective by using PAC-Bayes theory.

PAC-Bayes theory~\citep{McAllester1999COLT} enables us to analyze the generalization risk over a stochastic hypothesis class.
PAC-Bayes bounds of neural transfer learning~\citep{Galanti2016IMA, McNamara2017ICML} bound the generalization risk of a target task by the empirical risk of a source task.
In similar machine learning problems, studies on meta-learning~\citep{Amit2018ICML}, lifelong learning~\citep{Pentina2014ICML, Pentina2015NIPS}, and domain adaptation~\citep{Germain2016ICML}
also bound the generalization risk of a target task by the empirical risks of source tasks.
However, these bounds cannot directly be used for the simple heuristic sentence representation learning based on the publicly available pre-trained word vectors without their empirical risks.
In contrast, our transfer learning formulation and PAC-Bayes bound do not require the empirical risk of the source task.

Our contributions are as follows.
\begin{enumerate}
    \item We introduce the concept of generalization to sentence representation learning from pre-trained word vectors.
    \item We derive heuristic sentence vector methods from pre-trained word vectors by using PAC-Bayes theory.
    \item We propose novel unsupervised sentence representation learning on the basis of our PAC-Bayes analysis.
\end{enumerate}

\section{Preliminaries}
\subsection{Learning Word Vectors}
Word vector models aim to learn a map from each word to a $d$-dimensional real valued vector given a sequence of words $[w_1, \ldots, w_{T}]$.
This word vector is also called a \emph{word embedding} or a \emph{distributed word representation}.
In this paper, we use a popular neural word vector model, continuous skip-gram model, simply call Skip-gram, proposed by \citet{Mikolov2013ICLR}.
Skip-gram learns two types of vectors $\bfi$ and $\bfo$ associated with each word type.
We call $\bfi$ an input word vector and $\bfo$ an output word vector.
The terms $\hat{\bfi}$ and $\hat{\bfo}$ define a fixed input word vector and fixed output word vector trained on a sequence of words, respectively.

\subsubsection{Continuous Skip-gram with Negative Sampling \label{sec:skipgram}}
Intuitively, Skip-gram learns word vectors by predicting context word $w_c$ from given target word $w_t$ in a training sequence.
To do so, Skip-gram minimizes the negative log-likelihood of $p(w_c \mid w_t)$ modeled by a softmax function over training vocabulary $\calV$.
This straightforward training is intractable because evaluating the softmax layer takes $\calO(|\calV|)$.
A fast method for training Skip-gram model is negative sampling proposed by \citet{Mikolov2013NIPS}, whose loss function is defined by
\begin{align}
    L_{SG} &=
    \sum_{t=1}^T \sum_{w_c \in \mathcal{C}_t}  \ell_{neg}(\bfi_{w_t}, \bfo_{w_c}),
    \label{eq:sg-neg} \\
    \ell_{neg}(\bfi, \bfo) &= - \left[
        \ln \sigma \left( \bfi^\top \bfo \right)
        + \sum_{w_n \in \mathcal{NS}}  \ln \sigma \left( - \bfi^\top \bfo_{w_n} \right)
    \right],
    \label{eq:neg-loss}
\end{align}
where $\mathcal{C}_t$ represents a bag-of-words surrounding $w_t$ in a sequence,
$\sigma$ is a logistic sigmoid function,
and $\mathcal{NS}$ is a bag-of-words including $k$ negative words sampled from pre-defined noise distribution $p_n$.
Negative sampling loss \eqref{eq:neg-loss} can be considered as a new predictive task such that word vector $\bfi_{w_t}$ predicts whether word $w$ comes from context $\calC_t$ or noise distribution $p_n$ given $w_t$.

\subsection{Sentence Vectors from Pre-trained Word Vectors}
In the same way as word vector models,
sentence vector models aim to learn a map from a sentence\footnote{We call a sequence of words as a \textit{sentence} including a phrase, paragraph, and document.} to $d$-dimensional vector $\bfh_{\calS}$ given sentences $\{ \calS_1, \ldots, \calS_N \}$, where $\calS$ is a sequence of words.
Since sentences consist of words,
sentence representation is affected by word information, such as sentiment polarity.
We typically use pre-trained word vectors to learn sentence vectors because of the usefulness of their representations, which capture syntax and semantics~\citep{Mikolov2013NIPS}.
Given pre-trained word vectors $\{\hat{\bfi}_w, \hat{\bfo}_w : w \in \calV\}$,
the simplest way to obtain sentence vector ${\bfh}_{\calS}$ is to average the pre-trained word vectors of words appearing in sentence $\calS$,
for example,
\begin{align}
    \bfh_{\calS} &= \frac{1}{|\calS|} \sum_{w \in \calS} \hat{\bfi}_{w},
    \label{eq:average-sentence-vector}
    \\
    \bfh_{\calS} &= \frac{1}{|\calS|} \sum_{w \in \calS} \frac{\hat{\bfi}_{w} + \hat{\bfo}_{w}}{2}.
    \label{eq:both-average-sentence-vector}
\end{align}
Although these heuristics ignore the word order in a sentence,
empirically, they can outperform more complex deep neural network models in natural language processing tasks~\citep{Hill2016NAACL, Wieting2016ICLR, Arora2017ICLR}.

\subsection{PAC-Bayes Bound \label{sec:pac-bayes}}
We introduce a PAC-Bayes bound used in our analysis.
Let $\calD$ be an unknown test data distribution over $\calX \times \calY$,
where $\bfx \in \calX$ is an input data sample and $y \in \calY$ is an output data sample.
Let $\calD_S$ be a training dataset sampled i.i.d.\ from $\calD$, and $N$ be the number of training samples.
Let $\mathcal{P}$ be a prior distribution over hypotheses class $\calH$ and $\calQ$ be a posterior distribution over $\calH$.
Given the test data distribution and stochastic hypotheses, the generalization risk is defined as $R(\calQ) = \bbE_{h \sim \calQ} \bbE_{(\bfx, y) \sim \calD} \ell(\bfx, y, h)$,
where $\ell$ is a bounded loss function in $[0, 1]$.
\footnote{When the maximum of $\ell > 1$, we can use the same bound by using rescaling loss $\ell/\ell_{\max}$ in the same way as~\citet{Germain2016NIPS}, where $\ell_{\max}$ is the maximum value of $\ell$.}
In the same way, the empirical risk is defined as $\hat{R}(\calQ) = \bbE_{h \sim \calQ} \frac{1}{N} \sum_{i=1}^N \ell(\bfx_i, y_i, h)$.
Kullback-Leibler (KL) divergence is defined as $\KL(\calQ || \calP) = \bbE_{h \sim \calQ} \log \frac{\calQ(h)}{\calP(h)}$.
\begin{theorem}[PAC-Bayes Bound {\citep[Theorem 1.2.6]{Catoni2007}}\label{theorem:pac-bayes-bound}]
    $\forall \lambda > 0$, with probability at least $1-\delta$ over training samples $\calD_\calS$, $\forall \calQ$,
    \begin{align}
        R(\calQ) \leq
        \frac{
            1 - \exp \left[
                - \frac{\lambda}{N} \hat{R}(\calQ)
                - \frac{\KL(\calQ || \calP) - \log \delta}{N}
            \right]
        }{
            1 - \exp(-\frac{\lambda}{N})
        }.
        \label{eq:pac-bayes-bound}
    \end{align}
\end{theorem}
Intuitively, \cref{eq:pac-bayes-bound} means generalization risk $R(\calQ)$ is bounded by empirical risk $\hat{R}(\calQ)$ and hypothesis's complexity term $\KL(\calQ || \calP)$.
Hyperparameter $\lambda$ adjusts the trade-off between the empirical risk and the complexity term,
for example, the complexity term does not tend to contribute to the upper bound when $\lambda \to \infty$.

\section{PAC-Bayesian Analysis of Sentence Vectors \label{sec:pac-bayesian-anaysis-of-regression}}
We formulate learning word vectors and sentence vectors regarding transfer learning~\citep{Pan2010IEEE}.
In this section, we assume that the source task is Skip-gram with negative sampling.
Roughly, from our transfer learning perspective,
the source task minimizes the loss function of Skip-gram~\eqref{eq:sg-neg} on word sequences by updating input and output word vectors $\bfi$ and $\bfo$.
Then, a target task minimizes a loss function by updating sentence vector $\bfh_{\calS}$ with the fixed pre-trained word vectors $\hat{\bfi}$ and $\hat{\bfo}$ given sentences.
This transfer learning formulation enables us to analyze the generalization risk in learning sentence vectors with PAC-Bayes theory,
which can consider transferability to a target hypothesis from a learned source hypothesis through prior knowledge.
In \cref{sec:generalization}, we explain in more details our transfer learning setting and the concept of the generalization of learning sentence vectors,
and we also describe why the generalization becomes the matter to learn sentence vectors.

\subsection{Generalization and Predictive Sentence Vectors with Pre-trained Word Vectors \label{sec:generalization}}
Empirically, predictive word vector models, such as Skip-gram and CBoW~\citep{Mikolov2013ICLR, Mikolov2013NIPS},
stably outperform count models, such as a word co-occurrence matrix or its low-rank approximation with singular value decomposition in the natural language processing tasks~\citep{Baroni2014ACL-italic, Levy2015TACL}.
They learn word representations by solving a predictive task that makes learned word representations informative,
for example, Skip-gram with negative sampling~\eqref{eq:sg-neg} predicts whether word $w$ is from context $\calC_t$ or noise distribution $p_n$ by using target word's vector $\bfi_{w_t}$ described in~\cref{sec:skipgram}.
Predictive word vector models affect sentence vector models,
for example, the PV-DBOW model~\citep{Le2014ICML} predicts words in a paragraph given its paragraph id,
and the Skip-Thoughts model~\citep{Kiros2015NIPS} predicts the previous sentence and the next sentence given the current sentence.
Thus, it is essential for learning better sentence representations to solve predictive tasks.

We define test data distribution $\calD$ and training dataset $\calD_S$ to consider the generalization of predictive sentence vectors based on negative sampling loss for sentence $\calS$,
for example, ``\texttt{Paris is one of the most beautiful cities in the world}.''
The predictive task is to classify whether $w$ in $\calS$ comes from generative distribution or noise distribution $p_n$.
We assume that sentence $\calS$ is a bag-of-words,
and we also assume that each word $w$ in $\calS$ is sampled from a generative model based on a unigram language model with a Dirichlet prior:
\begin{align}
    & \boldsymbol{\phi} \sim \mathrm{Dirchlet}(\boldsymbol{\phi}; \boldsymbol{\gamma}), \\
    & w \sim \mathrm{Multinomial}(w; \boldsymbol{\phi}) \label{eq:multinomial-distribution},
\end{align}
where $\boldsymbol{\gamma} \in \mathbb{R}^{|\calV|}_{+}$ is a hyperparameter of the Dirichlet distribution.
We assume noise distribution $p_n$ is defined over all vocabulary, i.e., the unigram distribution,
and unknown test dataset $\calD$ generates word $w$ and label $y$ in the following ways:
\begin{align}
    y &\sim  \mathrm{Bernoulli}(y; \pi) = p^{\frac{y+1}{2}} (1-p)^{\frac{1-y}{2}} \text{ for } y \in \{-1, 1\},\\
    w &\sim  p(w \mid y; \boldsymbol{\phi}), \\
    p(w \mid y=1;\boldsymbol{\phi}) &= \mathrm{Multinomial}(w; \boldsymbol{\phi}), \\
    p(w \mid y=-1; \boldsymbol{\phi}) &= p_n(w),
\end{align}
where $\pi$ is a hyperparameter of the Bernoulli distribution.
Let an input data sample be $\bfx \in \mathcal{X} = \{\mathbf{x}_w \mid w \in \mathcal{V} \}$
and an output data sample be $y \in \calY = \{-1, 1\}$,
where $\bfx_w$ is a one-hot vector of word $w$, and $y$ is a binary label.
The observed training dataset denotes $\mathcal{D}_S = \{(\bfx_w, 1) \mid w \in \mathcal{S} \} \bigcup \{(\bfx_{w_i}, -1) \mid w_i \in p_n(w) \}_{i=1}^{k |\calS|}$,
where $k$ is the number of negative samples per each word in $\calS$.
This training dataset is sampled from $\calD$.
Test dataset $\calD$ contains positive data related to other words to describe \textit{Paris}, i.e., ``\texttt{France},'' ``\texttt{capital},'' and ``\texttt{incredible},'' in addition to the training dataset for the example sentence.

We formulate the binary classification problem as the problem mentioned earlier.
Let $h: \calX \to \calY$ be a hypothesis $h(\bfx ; \bfh_\calS, \mathbf{W}) = \mathrm{sign} [\bfh_{\calS}^{\top} (\bfW \bfx)]$ parameterized by sentence vector $\bfh_\calS$ and weight matrix $\bfW \in \bbR^{d \times |\calV|}$.
Note that $\bfh_\calS$ and $\bfW$ depend only on the sentence.
We expect $\bfh_\calS$ to capture a representation related to \textit{Paris} in the vector space $\bbR^d$ for correct classification.
Let loss function $\ell$ be zero-one loss $\ell(\bfx, y, h) = \bbI [h(\bfx; \bfh_\calS, \bfW) \neq y]$.
When we estimate $\bfh_\calS$ by minimizing usual empirical risk $L(\calD_S) = \frac{1}{|\calD_S| } \sum_{(x, y) \in \calD_S} \ell(\bfx, y, h)$,
$\bfh_\calS$ can capture empirical $\hat{\boldsymbol{\phi}}$ on the basis of the training data;
however, it does not become a feature representation such that the hypothesis predicts $y=1$ for a word excluding $\calD_S$ but it is likely to be sampled from~\cref{eq:multinomial-distribution}, i.e., ``\texttt{France}.''
As the result of minimizing the empirical risk, this sentence vector is poor to represent the sentence's meaning to describe \textit{Paris},
in other words, the sentence vector overfits on the training dataset.
This is why we need to consider the generalization of learning sentence vectors.
Unfortunately, we cannot minimize the generalization risk, $L(\calD) = \bbE_{(x, y) \sim \calD} \ell(\bfx, y, h)$, since this risk depends on unknown $\calD$.
Therefore we need to minimize an upper bound of the generalization risk.

We use the PAC-Bayes framework to bound generalization risk $R$ because it is empirically tighter than Vapnik-Chervonenkis dimension based generalization bounds~\citep{Dziugaite2017UAI, McNamara2017ICML}.
We apply the PAC-Bayes framework to the previous predictive task.
PAC-Bayes theory assumes $h$ is a stochastic hypothesis.
That means sentence vector $\bfh_\calS$ is sampled from posterior $\calQ$ such as multivariate Gaussian distribution $\calN(;\boldsymbol{\mu}_\calS, \sigma_{\calQ}^2 I)$,
where $\boldsymbol{\mu}_\calS$ is a $d$ dimensional mean vector, $\sigma_{\calQ}^2$ is a variance parameter, and $I$ is the identity matrix.
We can also introduce a posterior distribution related to $\bfW$.
However, on the basis of empirical studies~\citep{Baroni2014ACL-italic, Levy2015TACL},
we use fixed Skip-gram's pre-trained vectors $\{\hat{\bfi}_w \mid w \in \calV \} $ as $\bfW$ where the $w$-th column of $\bfW$ corresponds to $\hat{\bfi}_w$,
because these vectors already capture considerable word syntax and semantics on the source task's sequence.
That means $\bfW$ is not sampled from a posterior distribution over the hypothesis class in the PAC-Bayesian framework.

By using pre-trained word vectors in the hypothesis,
we can formulate transfer learning where the source task is learning word vectors by Skip-gram with negative sampling on a large unlabeled corpus,
and the target task is learning posterior parameters $\boldsymbol{\mu}_\calS$ and $\sigma^2_{\calQ}$ with fixed pre-trained word vectors by minimizing \cref{eq:pac-bayes-bound} including the predictive task.

In this transfer learning formulation and PAC-Bayes framework,
it is natural to use pre-trained output word vectors $\{\hat{\bfo}_w \mid w \in \calV \}$ as the parameters of prior distribution $\calP$ of $\mathbf{h}_\calS$
because words in the sentence probably co-occur in the source task's sequence as well,
i.e., ``\texttt{Paris}'' and ``\texttt{beautiful}.''
Similar to the posterior, we use a multivariate Gaussian distribution parameterized by output word vectors,
for example, the summing over vectors of words appearning in $\calS$, $\mathcal{P} = \mathcal{N}(; \sum_{w \in \calS} \hat{\bfo}_w ,\sigma^2_{\calP} I)$,
where $\sigma^2_\calP$ is a variance parameter.
The prior prevents the posterior from overfitting on the target's training data by penalizing the posterior far from the prior via KL divergence in \cref{eq:pac-bayes-bound}.

\subsection{Loss Function for Sentence Vector Training Based on PAC-Bayes Bound}
We define a PAC-Bayes bound to analyze sentence representation learning defined by \cref{eq:average-sentence-vector} and \cref{eq:both-average-sentence-vector} on the basis of the setting described in \cref{sec:generalization}.
Given sentence $\calS$ and pre-trained input and output word vectors in the source task,
let $|\calS|$ be the number of words in $\calS$ and $\calD_S$ be a training dataset sampled i.i.d.\ from $\calD$.
We define prior distribution over hypothesis class $\calH$ as $\calP_{\calS} \propto \prod_{w \in \calS} \calN(; \hat{\bfo}_w, \sigma_{\calP}^2 I)$.
This prior distribution is known as the product of experts~\citep{Hinton2002NC} such that each expert is a multivariate Gaussian distribution parameterized by $\hat{\bfo}$ of the words appearing in $\calS$.
We also define a posterior distribution as $\calQ_{\calS} := \calN(; \boldsymbol{\mu}_{\calS}, \sigma_{\calQ}^2 I)$.
\cref{theorem:loss-Transfer-vectors} shows the loss function based on PAC-Bayes bound.
\begin{theorem}[Loss Function of Learning Sentence Vector from Pre-trained Word Vectors Based on PAC-Bayes Bound\label{theorem:loss-Transfer-vectors}]
    Let $\calQ_{\calS} = \calN(; \boldsymbol{\mu}_\calS, \sigma_{\calQ}^2 I)$ be the posterior of sentence vector $\bfh_\calS$.
    Given $\lambda > 0$, $\calS$, $\calD_S$,
    $\calP_{\calS} \propto \prod_{w \in \calS} \calN(; \hat{\bfo}_w, \sigma_{\calP}^2 I)$,
    and bounded loss $\ell$,
    minimizing \cref{eq:pac-bayes-bound} is equivalent to
    minimizing the loss function defined by
    \begin{align}
        L(\mathcal{D}_S, \calQ_\calS) =
            \bbE_{h \sim \calQ_\calS} \frac{1}{N} \sum_{i=1}^{N} \ell(\bfx_i, y_i, h)
            + \frac{|\calS|}{2 \sigma^2_\calP \lambda} \left|\left| \boldsymbol{\mu}_{\calS} - \frac{1}{|\calS|} \sum_{w \in \calS} \hat{\bfo}_w \right|\right|_2^2
            + \frac{|\calS| d}{2 \lambda}
                \left(
                    \ln \frac{\sigma^2_\calP}{\sigma^2_\calQ} + \frac{\sigma_\calQ^2}{\sigma_\calP^2}
                \right)
                + C
            \label{eq:general_loss}
    \end{align}
    where $C$ is a constant term that does not depend on $\calQ$.
\end{theorem}
We prove \cref{theorem:loss-Transfer-vectors} by using~\cref{theorem:pac-bayes-bound}.
The proof is shown in~\cref{sec:proof-bayes-bound}.
In \cref{eq:general_loss}, the first term is the empirical risk of a target task,
and the second and the third terms penalize posterior $\calQ_\calS$ far from prior $\calP_\calS$.

\subsection{Sentence Vectors with Squared L2 Loss}
We use \cref{theorem:loss-Transfer-vectors} to analyze heuristic sentence vector models.
Intuitively, this predictive task is to predict a pre-trained word vector from a hypothesis directly.
Let $\calX := \{ \bfx_w \mid w \in \calV \}$, where $\bfx_w$ represents a one-hot vector of word $w$.
Let $\calY := \{\hat{\bfi}_w \mid w \in \calV\}$ and
hypothesis $h$ be $h(\bfx) := \bfh_\calS$.
Let training samples $\calD_S := \{(\bfx_w, \hat{\bfi}_w) \mid w \in \calS\}$.
Each sentence does not depend on other sentences at all,
so hypothesis $h$ always returns sentence vector $\bfh_\calS$.
We use squared L2 loss function $\ell(\bfx, y, h) := \frac{1}{2} || y - h ||_2^2$.
We assume that $\ell$ is bounded in $[0, \ell_{\max}]$.
In this setting,
we can obtain the closed form of parameters of $\calQ_\calS$:
\begin{align}
    \boldsymbol{\mu}_{\calS} =& \frac{1}{ (1 + \alpha) |\calS|} \sum_{w \in \calS} \left( \hat{\bfi}_{w} + \alpha \hat{\bfo}_w \right) \label{eq:avg}, \\
    \sigma^2_\calQ =& \sigma^2_\calP \label{eq:variance-avg},
\end{align}
where $\alpha = \frac{|\calS|}{\sigma_\calP^2 \lambda}$.
Details on solutions are given in \cref{sec:losses}.
From~\cref{eq:avg}, we can obtain two heuristic sentence representation learning algorithms.
\begin{corollary}[Sentence Vector by Averaging Pre-trained Input Word Vectors\label{cor:average}]
    Posterior's mean vector $\boldsymbol{\mu}_{\calS}$ that is estimated by minimizing~\cref{eq:general_loss} with $\alpha =0$,
    in other words, $\lambda \to \infty$,
    is equivalent to~\cref{eq:average-sentence-vector}, $\frac{1}{|\calS|} \sum_{w \in \calS} \hat{\bfi}_{w}$.
\end{corollary}

\begin{corollary}[Sentence Vector by Averaging Pre-trained Input and Output Word Vectors\label{cor:average-both}]
    Posterior's mean vector $\boldsymbol{\mu}_{\calS}$ that is estimated by minimizing~\cref{eq:general_loss} with $\alpha = 1$,
    is equivalent to~\cref{eq:both-average-sentence-vector}, $\frac{1}{|\calS|} \sum_{w \in \calS} \frac{\hat{\bfi}_{w} + \hat{\bfo}_{w}}{2}$.
\end{corollary}
Empirically, the averaged vector of an input word vector and an output word vector,
$\frac{\hat{\bfi} + \hat{\bfo}}{2}$, can improve the performance of downstream tasks, e.g.,~\citet{Levy2015TACL}.
\cref{cor:average-both} offers a novel perspective on the operation in the heuristic sentence representation learning algorithm.

\subsubsection{IDF Weighting}
Sentence vectors by averaging weighed word vectors with inverse document frequency (IDF) is another simple heuristic~\citep{Lilleberg2015ICCICC}.
IDF is also a widely used heuristic to weight a word in text collections.
The IDF of word $w$ is defined by
\begin{align}
    \mathrm{IDF}(w) = \log \left(
        \frac{N_{\calS}}{\sum_{i=1}^{N_{\calS}} \bbI \left[w \in \calS_i \right]}
        \right) + 1,
    \label{eq:idf}
\end{align}
where $N_\calS$ is the number of training sentences.
Our interest is to derive this heuristic from our PAC-Bayes bound.
We change the loss function to sample-dependent weighted loss function $\ell(\bfx, y, h) = \beta(\bfx)||y - h(\bfx)||_2^2$,
where $\beta$ is a weighting function from a one-hot vector to $\bbR_+$.
We also change the prior to $\calP_{\calS} \propto \prod_{w \in \calS} \calN(; \hat{\bfo}_w, \sigma^2_{\calP_w} I)$.
When we set $\beta(\bfx) = |\calS| \mathrm{IDF}(\bfx)$ and $\sigma^2_{\calP_w} = \mathrm{IDF}(w)^{-1}$,
we can obtain the closed form of $\boldsymbol{\mu}_\calS$ and $\sigma^2_\calQ$:
\begin{align}
    \boldsymbol{\mu}_\calS =& \frac{1}{\left(1 + \frac{1}{\lambda} \right) \sum_{w \in \calS} \mathrm{IDF}(w)} \sum_{w \in \calS} \mathrm{IDF}(w) \left(\hat{\bfi}_{w} + \frac{1}{\lambda} \hat{\bfo}_w \right) \label{eq:idf-weighting}, \\
    \sigma^2_\calQ =& \left( \frac{1}{|\calS|} \sum_{w \in \calS} \frac{1}{\sigma^2_{\calP_w}} \right)^{-1}. \label{eq:idf-variance-avg}
\end{align}
The details are in~\cref{sec:idf}.
This sentence vector differs slightly from IDF weighting of a sentence vector proposed by~\citet{Lilleberg2015ICCICC} since word vectors are weighted relatively by IDF in a sentence.

\subsection{Sentence Vectors with Zero-one Loss}
From our PAC-Bayes analysis of simple heuristics, we found that the problem, i.e., the loss function, differs between the source and target tasks in existing heuristics.
From a transfer learning perspective, a target task's loss value tends to decrease easily if the source and target tasks are similar, i.e., loss function, data distribution, and task.
The new target task is based on the problem setting in \cref{sec:generalization}: predicting whether a word in a sentence comes from a generative distribution or a noise distribution.
We expect that sentence vectors are similar in embedded space if sentences share similar meanings when the generalization risk is sufficiently small.

We apply \cref{theorem:loss-Transfer-vectors} to the predictive task defined in \cref{sec:generalization}.
Let the output space be $\calY := \{-1, 1\}$,
loss function $l$ be zero-one loss $\ell(\bfx, y, h) := \bbI[h(\bfx) \neq y]$,
and the hypothesis be $h(\bfx_w) := \sign[\bfh_{\calS}^\top (\hat{\bfI} \bfx_w)]$,
where the $w$-th column of $\hat{\bfI} \in \bbR^{d \times |\calV|}$ corresponds to the $w$-th pre-trained input word vector $\hat{\bfi}_{w}$.
Hypothesis distributions $\calP$ and $\calQ$ are the same as in the squared L2 loss case.
In practice, we minimize the following loss function by using negative sampling as a surrogate loss function of zero-one loss:
\begin{align}
    L(\calD_S, \calQ_\calS)
    = \bbE_{h \sim \calQ_\calS} \frac{1}{|\calS|}
        \sum_{w \in \calS}
        \ell_{neg}(\bfh_{\calS}, \hat{\bfi}_w)
    + \frac{|\calS|}{2 \sigma_\calP^2 \lambda}
        \left|\left| \boldsymbol{\mu}_{\calS} - \frac{1}{|\calS|} \sum_{w \in \calS} \hat{\bfo}_w \right|\right|_2^2
    + \frac{|\calS| d }{2 \lambda}
        \left(
            \ln \frac{\sigma^2_\calP}{\sigma^2_\calQ} + \frac{\sigma_\calQ^2}{\sigma_\calP^2}
        \right).
        \label{eq:proposed-loss}
\end{align}
\begin{corollary}[Relationship to Paragraph Vector Models]
    PV-DBOW, an instance of paragraph vector models proposed by~\citet{Le2014ICML}, with negative sampling loss is the same as~\cref{eq:proposed-loss} with $\lambda \to \infty$ and with trainable word vector $\bfi$ from scratch and do not take the expectation over the posterior.
\end{corollary}

\section{Other Transfer Learning Settings \label{sec:other-transfers}}
\subsection{Another Sentence Vector Modeling for Target Task: Learnable Word Vectors}\label{sec:words-based-sentence-vectors}
Minimizing~\cref{eq:proposed-loss}
is inefficient because each sentence vector's posterior $\calQ$ is completely independent of other sentences.
We consider another sentence vector's modeling based on trainable word vectors defined by
\begin{align}
    \bfh_{\calS} = \frac{1}{|\calS|} \sum_{w \in \calS} \bfh_{w}.
    \label{eq:averaging-model}
\end{align}
We estimate posterior's parameters of word vector $\bfh_w$ on the target task instead of estimating posterior's parameters of $\bfh_\calS$ directly.
This formulation is similar to supervised \texttt{fastText}~\citep{EACL2017Joulin} and \texttt{fastSent}~\citep{Hill2016NAACL}.
Thanks to this word-based modeling, we do not need to learn the vectors of new sentences.
Moreover, we can initialize each word vector $\boldsymbol{\mu}_w$ by the prior's pre-trained word vector $\hat{\bfo}$.
Let $\calX := \{ (\calS_i, \bfx_w) \mid i \in \bbN_{\leq N_{\calS}} , w \in \calV \}$,
the prior of $\bfh_w$ be $\calP_w = \calN(\hat{\bfo}_w, \sigma_{\calP_w}^2 I)$,
and let the posterior of hypothesis be $\calQ_w = \calN(; \boldsymbol{\mu}_w, \sigma_{\calQ_w}^2 I)$ for all $w$.
\begin{theorem}[Loss Function of Word Based Learning Sentence Vector from Pre-trained Word Vectors Based on PAC-Bayes Bound\label{theorem:word-based-loss-transfer-vectors}]
    Let $\calQ = \{\calN(; \boldsymbol{\mu}_w, \sigma_{\calQ_w}^2 I) \mid w \in \calV\}$ be the posterior of word vector $\bfh_w$.
    Given $\lambda > 0$,
    $\calD_S$,
    $\calP_{w} = \{\calN(; \hat{\bfo}_w, \sigma_{\calP_w}^2 I) \mid w \in \calV \}$,
    and bounded loss $\ell$,
    minimizing \cref{eq:pac-bayes-bound} is equivalent to
    minimizing the loss function defined by
    \begin{align}
        L(\mathcal{D}_S, \calQ) = \bbE_{h \sim \calQ} \frac{1}{N} \sum_{i=1}^{N} \ell(\bfx_i, y_i, h)
        + \frac{1}{2 \lambda} \sum_{w \in \calV} \left[
                \frac{1}{\sigma_{\calP_w}^2} \left|\left| \boldsymbol{\mu}_w - \hat{\bfo}_w \right|\right|_2^2
            + d \left(
                \ln \frac{\sigma^2_{\calP_w}}{\sigma^2_{\calQ_w}} + \frac{\sigma_{\calQ_w}^2}{\sigma_{\calP_w}^2}
            \right)
        \right] + C,
            \label{eq:word_general_loss}
    \end{align}
    where $C$ is a constant term that does not depend on posterior parameters.
\end{theorem}
The proof is almost the same as \cref{theorem:loss-Transfer-vectors}.
The difference is replacing the KL term with $\KL(\calQ || \calP) = \\ \sum_{w \in \calV} \KL(\calQ_w || \calP_w)$.

The practical surrogate loss function using negative sampling loss defined by
\begin{align}
L(\calD_S, \calQ) = \bbE_{h \sim \calQ} \frac{1}{N} \sum_{(\calS, w) \sim \calD_S} \ell_{neg} \left( \frac{1}{|\calS|} \sum_{v \in \calS} \bfh_v, \hat{\bfi}_w \right)
    + \frac{1}{2 \lambda} \sum_{w \in \calV} \left[
            \frac{1}{\sigma_{\calP_w}^2} \left|\left| \boldsymbol{\mu}_w - \hat{\bfo}_w \right|\right|_2^2
    + d \left( \ln \frac{\sigma^2_{\calP_w}}{\sigma^2_{\calQ_w}} + \frac{\sigma_{\calQ_w}^2}{\sigma_{\calP_w}^2} \right)
        \right].
    \label{eq:proposed-loss-sentence}
\end{align}
Recall that our goal is to obtain general sentence representations.
However, \cref{eq:proposed-loss-sentence} means that lower frequency words do not relatively change from the prior because negative sampling loss depends on the word frequency when $\frac{1}{\sigma_{\calP_w}^2}$ is the same for all words.
In natural language data, high-frequency words are meaningless to represent sentence meaning because they appear among almost all sentences, for example, ``\texttt{the}'' and ``\texttt{to}.''
To avoid learning such poor sentence representations,
we define the variance parameters of each prior distribution as
\begin{align}
    \sigma^2_{\calP_w} = \frac{\sum_{v \in \calV} freq(v)}{freq(w)},
    \label{eq:word-depend-variance}
\end{align}
where $freq(w)$ is the frequency of word $w$ in a training dataset.
Intuitively, this prior penalizes higher frequency words rather than lower frequency words.

\subsection{Switching Pre-trained Word Vectors' Role \label{sec:swap-prior}}
We consider another target task setting, that is switching the role of fixed input word vectors and fixed output word vectors.
For squared L2 loss and zero-one loss,  we use fixed pre-trained output word vectors in output samples and the hypothesis respectively,
and we use fixed pre-trained input word vectors as the prior distribution's mean vector.

\begin{table*}[tb]
    \caption{
        Comparison of source tasks and target tasks in our transfer learning.
        \label{tb:comparison-tasks}
    }
    \centering
    \begin{tabular}{cccccc}
\toprule
                Model &       Input $\mathbf{x}$ &            Output $y$ &                                                                                                  Hypothesis $h$ &          Loss $\ell$ & Prior's word vectors \\
\midrule
            Skip-gram &               $w_t, w_c$ &                Binary &                                                             $\sigma(\mathbf{i}^{\top}_{w_t} \mathbf{o}_{w_c} )$ &    Negative sampling &                   Negative sampling \\ 
 \midrule 

       \texttt{PB-L2} &           $\mathbf{x}_w$ &  $\hat{\mathbf{i}}_w$ &                                                                                            $\mathbf{h}_{\calS}$ &           Squared L2 &   $\hat{\mathbf{O}}$ \\
     \texttt{i-PB-L2} &           $\mathbf{x}_w$ &  $\hat{\mathbf{o}}_w$ &                                                                                            $\mathbf{h}_{\calS}$ &           Squared L2 &   $\hat{\mathbf{I}}$ \\
   \texttt{PB-IDF-L2} &           $\mathbf{x}_w$ &  $\hat{\mathbf{i}}_w$ &                                                                                            $\mathbf{h}_{\calS}$ &  Weighted squared L2 &   $\hat{\mathbf{O}}$ \\
 \texttt{i-PB-IDF-L2} &           $\mathbf{x}_w$ &  $\hat{\mathbf{o}}_w$ &                                                                                            $\mathbf{h}_{\calS}$ &  Weighted squared L2 &   $\hat{\mathbf{I}}$ \\
      \texttt{PB-neg} &  $(\calS, \mathbf{x}_w)$ &                Binary &                                         $\sigma[\mathbf{h}_{\mathcal{S}}^\top (\hat{\mathbf{I}} \mathbf{x}_w)]$ &    Negative sampling &   $\hat{\mathbf{O}}$ \\
    \texttt{i-PB-neg} &  $(\calS, \mathbf{x}_w)$ &                Binary &                                         $\sigma[\mathbf{h}_{\mathcal{S}}^\top (\hat{\mathbf{O}} \mathbf{x}_w)]$ &    Negative sampling &   $\hat{\mathbf{I}}$ \\
    \texttt{w-PB-neg} &  $(\calS, \mathbf{x}_w)$ &                Binary &  $\sigma[(\frac{1}{|\mathcal{S}|} \sum_{w \in \mathcal{S}} \mathbf{h}_w)^\top (\hat{\mathbf{I}} \mathbf{x}_w)]$ &    Negative sampling &   $\hat{\mathbf{O}}$ \\
  \texttt{i-w-PB-neg} &  $(\calS, \mathbf{x}_w)$ &                Binary &  $\sigma[(\frac{1}{|\mathcal{S}|} \sum_{w \in \mathcal{S}} \mathbf{h}_w)^\top (\hat{\mathbf{O}} \mathbf{x}_w)]$ &    Negative sampling &   $\hat{\mathbf{I}}$ \\
\bottomrule
\end{tabular}

\end{table*}
\cref{tb:comparison-tasks} summarizes all source and target tasks discussed in this paper.
We call sentence vector \eqref{eq:avg} as \texttt{PB-L2}
and \eqref{eq:idf-weighting} as \texttt{PB-IDF-L2}.
We also call sentence vectors minimizing~\cref{eq:proposed-loss} as \texttt{PB-neg},
and we call sentence vectors minimizing~\cref{eq:proposed-loss-sentence} with~\cref{eq:word-depend-variance} as \texttt{w-PB-neg}.
We add the prefix ``\texttt{i-}'' to their names when we switch the roles of input words vectors and output word vectors.

\section{Experiments}
We verified our analysis in sentence classification tasks because learned sentence vectors work as feature vectors for supervised machine learning tasks.
We compared simple heuristic methods derived by our analysis and our sentence vector learning methods.
We trained Skip-gram on a large text corpus as a source task.
Using pre-trained vectors of Skip-gram,
we estimated posterior's parameters, $\boldsymbol{\mu}$ and $\sigma^2_{\calQ}$, obtained from our analysis as target tasks, which are shown in~\cref{tb:comparison-tasks}.
For \texttt{w-PB-neg} and \texttt{i-w-PB-neg}, we calculated a sentence vector of $\calS$ by averaging over posterior's $\boldsymbol{\mu}_w$ of $w$ appearing in $S$.
We used posterior's $\boldsymbol{\mu}_\calS$ as a sentence vector of $\calS$ for other models.

\subsection{Settings of the Source Task}
We used English Wikipedia articles\footnote{We downloaded XML dump file created on Aug. 1, 2018.} to train word vectors.
We pre-processed this corpus with \texttt{wikifil.pl}\footnote{\url{https://github.com/facebookresearch/fastText/blob/master/wikifil.pl}}, and
then removed words appearing less than $5$ times.
In the pre-processed corpus, the size of vocabulary $|\mathcal{V}|$ was $2\ 472\ 211$, and the number of tokens was $4\ 517\ 598\ 626$.

We used the following hyperparameters to train word vectors with \texttt{word2vec}\footnote{\url{https://github.com/nzw0301/word2vec}}:
word vector dimensionality $d$ was $300$,
the window size was $5$,
the sub-sampling parameter was $0.0001$,
the number of iterations was $5$,
the number of negative samples $k$ was $15$,
the noise distribution parameter was $\frac{3}{4}$, and
the initial learning rate was $0.025$.

\subsection{Settings of Target Tasks}
\subsubsection{Classification Datasets}
\begin{table*}[tb]
    \caption{
        Sentence Classification Datasets
        \label{tb:supervised_datasets}
    }
    \centering
    \begin{tabular}{ccrrr}
        \toprule
        Dataset                     & Task                        & \#Train data  & \#Test data & \#Classes \\
        \midrule
        20news~\citep{Lang1995ICML} & Topic classification        & $11\,314$     &    $7\,532$ & $20$ \\
        IMDb~\citep{Maas2011ACL}    & Sentiment analysis          & $25\,000$     &   $25\,000$ &  $2$ \\
        SUBJ~\citep{Pang2004ACL}    & Subjectivity classification &  $8\,000$     &    $2\,000$ &  $2$ \\
        \bottomrule
    \end{tabular}
\end{table*}
We used three classification datasets for the target tasks to learn sentence vectors:
1) 20 news topic classification (20news~\citep{Lang1995ICML})
2) movie review's sentiment analysis (IMDb~\citep{Maas2011ACL}),
and 3) movie review's subjectivity classification (SUBJ~\citep{Pang2004ACL}).
\cref{tb:supervised_datasets} shows the three classification datasets.
We pre-processed both datasets in the same manner as pre-processing of the Wikipedia corpus by using \texttt{wikifil.pl}.
We split the SUBJ dataset randomly into 80 \% as training data and 20 \% as test data for evaluation because this dataset was not split into training data and test data.
Note that we \textit{did not use} label information at all while training target tasks.

\subsubsection{Settings of Sentence Vector Methods}
We trained simple heuristic algorithms derived by our analysis.
We called~\cref{eq:avg} with $\alpha \in\{0, 1\}$ as \texttt{Average},
and we call~\cref{eq:idf-weighting} with $\frac{1}{\lambda} \in \{0, 1\}$ as \texttt{IDF-Average}.
In the same way as the notation on~\cref{tb:comparison-tasks}, the prefix ``\texttt{i-}'' means that we switched the roles of input words vectors and output word vectors.
Note that sentence vectors of \texttt{Average} with $\alpha=1$ and \texttt{i-Average} with $\alpha=1$ are the same,
and also sentence vectors of \texttt{IDF-Average} with $\frac{1}{\lambda}=1$ and \texttt{i-IDF-Average} with $\frac{1}{\lambda}=1$ are the same.
So we omitted the results of \texttt{i-Average} with $\alpha=1$ and \texttt{i-IDF-Average} with $\frac{1}{\lambda}=1$.

For the squared L2 loss based models, \texttt{PB-L2}, \texttt{PB-IDF-L2}, \texttt{i-PB-L2}, and \texttt{i-PB-IDF-L2},
we trained the sentence vectors with $\lambda \in \{0.25, 0.5, 1, 2, 4, 8\}$ and fixed the prior's variance parameter $\sigma^2_\calP = \sigma^2_{\calP_w} = 1$.
Hyperparameters $\lambda$ was searched by grid-search in the supervised tasks.

For the zero-one loss based models, \texttt{PB-neg}, \texttt{w-PB-neg}, \texttt{i-PB-neg}, and \texttt{i-w-PB-neg},
we used the following parameters to train sentence vectors:
the number of iterations was $40$,
the number of negative samples $k$ was $15$,
noise samples were the uni-gram distribution of target task's training sentences powered by $\frac{3}{4}$,
initial learning rate $\eta$ was the same as Skip-gram, $0.025$.
We used stochastic gradient descent to optimize sentence vectors with mini-batch as a sentence with linearly decreased the learning rate per epoch.
We fixed $\sigma_\calP^2 = 1$ for \texttt{PB-neg} and \texttt{i-PB-neg}.
Hyperparameter $\lambda$ was searched in $\{0.25, 0.5, 1, 2, 4, 8\}$ by cross-validation of the classification phase.
Estimated parameters of \texttt{PB-neg} and \texttt{i-PB-neg}, $\boldsymbol{\mu}_\calS$ and $\sigma^2_{\calS}$, were initialized by $U(-\frac{0.5}{300}, \frac{0.5}{300})$,
and estimated parameters of \texttt{w-PB-neg} and \texttt{i-w-PB-neg} were initialized by prior's parameters.
We implemented all algorithms with \texttt{PyTorch}~\citep{paszke2017automatic}.
We optimized the loss functions by using reparameterization trick~\citep{Kingma2014ICLR} for multivariate Gaussian distribution to take the expectation of posterior $\calQ$.
As an implementation technique to accelerate training,
we applied lazy stochastic gradient descent~\citep{Carpenter2008techRepo} to optimize~\cref{eq:proposed-loss-sentence}.
We directly updated parameters not using automatic differentiation because we needed to evaluate trainable word vectors for each update.

\subsection{Evaluation of Sentence Representations}
We evaluated sentence vectors trained on the target tasks as feature vectors of classification tasks.
We implemented a one-vs-rest logistic regression classifier with \texttt{scikit-learn}~\citep{scikit-learn} and \texttt{GNU Parallel}~\cite{tange_ole_2018_1146014}.
We chose hyperparameter $C$ of logistic regression and the pre-processing parameters as L2 normalization of sentence vectors on the basis of grid-search with five-fold cross-validation.
All classification scores were values averaged over three times with different random seeds.

\subsection{Classification Results}
\begin{table*}[tb]
    \caption{
        Test accuracy of sentence classification (averaged over three times).
        Note that rows of \texttt{i-Average} with $\alpha=1$ and \texttt{i-IDF-Average} with $\frac{1}{\lambda}=1$ were omitted because their results were the same as the result without \texttt{i-} models.
        \label{tb:classification-results}
    }
    \centering
    \begin{tabular}{ccccl}
\toprule
                                          Model & &            20news &               IMDb &               SUBJ \\
\midrule
                   \texttt{Average} &$\alpha=0$ &  0.748 $\pm$ 0.000 &  0.842 $\pm$ 0.001 &  \textbf{0.908 $\pm$ 0.001} \\
                   \texttt{Average} &$\alpha=1$ &  0.745 $\pm$ 0.001 &  0.838 $\pm$ 0.000 &  0.904 $\pm$ 0.000 \\
   \texttt{IDF-Average} &$\frac{1}{\lambda}=0$ &  0.737 $\pm$ 0.001 &  0.823 $\pm$ 0.001 &  0.907 $\pm$ 0.000 \\
   \texttt{IDF-Average} &$\frac{1}{\lambda}=1$ &  0.735 $\pm$ 0.000 &  0.821 $\pm$ 0.000 &  0.906 $\pm$ 0.001 \\
                 \texttt{i-Average} &$\alpha=0$ &  0.752 $\pm$ 0.000 &  0.842 $\pm$ 0.000 &  0.906 $\pm$ 0.001 \\
 \texttt{i-IDF-Average} &$\frac{1}{\lambda}=0$ &  0.737 $\pm$ 0.000 &  0.822 $\pm$ 0.000 &  0.902 $\pm$ 0.000 \\
 \midrule
                                \texttt{PB-L2}  &&  \textbf{0.753 $\pm$ 0.000} &  0.841 $\pm$ 0.000 &  0.905 $\pm$ 0.000 \\
                            \texttt{PB-IDF-L2}  &&  0.735 $\pm$ 0.002 &  0.823 $\pm$ 0.000 &  0.907 $\pm$ 0.000 \\
                              \texttt{i-PB-L2}  &&  0.748 $\pm$ 0.001 &  0.841 $\pm$ 0.000 &  0.908 $\pm$ 0.000 \\
                          \texttt{i-PB-IDF-L2}  &&  0.737 $\pm$ 0.001 &  0.823 $\pm$ 0.000 &  0.906 $\pm$ 0.000 \\
                                \texttt{PB-neg} &&  0.750 $\pm$ 0.000 &  0.842 $\pm$ 0.000 &  0.907 $\pm$ 0.000 \\
                              \texttt{w-PB-neg} &&  0.752 $\pm$ 0.000 &  0.843 $\pm$ 0.000 &  0.902 $\pm$ 0.001 \\
                              \texttt{i-PB-neg} &&  0.748 $\pm$ 0.001 &  0.842 $\pm$ 0.000 &  \textbf{0.910 $\pm$ 0.001} \\
                              \texttt{i-w-PB-neg} &&  0.750 $\pm$ 0.001 &  \textbf{0.844 $\pm$ 0.000} &  \textbf{0.910 $\pm$ 0.001} \\
\bottomrule
\end{tabular}

\end{table*}
\cref{tb:classification-results} shows the test accuracies and their standard deviation of the sentence classification results.
Test accuracies were slightly different among models.
Zero-one loss based models performed more stably than squared L2 loss based models and heuristics did on 20news and IMDb datasets.

\section{Conclusion}
We formulated learning sentence vectors from neural word vector models as transfer learning.
We derived heuristic sentence vector models by applying PAC-Bayes theory to target tasks.
Our analysis of sentence vectors is a first step towards understanding of practical sentence vector representation learning.
We also proposed novel sentence representation learning on the basis of our PAC-Bayes analysis by replacing the loss function and hypothesis class.
In our experiments, the performance of all sentence vector models was almost the same in sentence classification.

Recently, sentence representations transferred by bi-directional recurrent neural networks language modeling
have archived the state-of-the-art performance in natural language processing tasks~\citep{Howard2018ACL, Peters2017ACL}.
Analyzing these complex deep predictive representation models by PAC-Bayes theory is left to future work.
We also would like to consider a technique of Skip-gram training, sub-sampling proposed by~\citet{Mikolov2013NIPS}, to make the target task more similar to the source task.

\newpage

\section*{Acknowledgements}
We thank Ikko Yamane, Hideaki Imamura, Makoto Hiramatsu, Seiichi Kuroki, and Futoshi Futami for useful discussions and their helpful comments.
We also thank developers of \texttt{scikit-learn}, \texttt{gensim}, and \texttt{PyTorch}.
KN was supported by JSPS KAKENHI Grant Number 18J20470.
IS was supported by JSPS KAKENHI Grant Number 17H04693.

\newpage

\appendix
\section{Proposed PAC-Bayes Bound and Concrete Loss Functions}
\subsection{Proof of~\cref{theorem:loss-Transfer-vectors} \label{sec:proof-bayes-bound}}
\begin{proof}
Given sentence $\calS$ and bounded loss $\ell$,
$\calP_{\calS} = \frac{\prod_{w \in \calS} \calN(; \hat{\bfo}_{w}, \sigma^2_\calP I)}{\int \prod_{w \in \calS} \calN(\bfh; \hat{\bfo}_{w}, \sigma^2_\calP I) \idh}$,
and $\forall \lambda > 0$ with probability at least $1-\delta$ over training samples $\calD_S$,
minimization of~\cref{eq:pac-bayes-bound} is equivalent to minimizing the loss function defined by
\begin{align}
    L(\calD_S, \calQ) = \hat{R}(\calQ_\calS) + \frac{1}{\lambda} \KL (\calQ_\calS || \calP_\calS).
    \label{eq:pac-bayes-loss}
\end{align}
We focus on $\KL(\calQ_{\calS} || \calP_{\calS})$ term in~\cref{eq:pac-bayes-loss}.
Recall that posterior $\calQ_{\calS} = \calN(;\boldsymbol{\mu}_\calS, \sigma^2_\calQ I)$.
\begin{align}
    \KL(\calQ_{\calS} || \calP_{\calS})
    &= \int \calQ_{\calS} \log \frac{\calQ_\calS}{\calP_\calS} \idh \\
    &= \int \calQ_{\calS} \log \frac{\calQ_\calS}{
        \left[
            \frac{
                \prod_{w \in \calS} \calN(; \hat{\bfo}_{w}, \sigma^2_\calP I)
            }{
                \int \prod_{w \in \calS} \calN(\bfh; \hat{\bfo}_{w}, \sigma^2_\calP I) \idh
            }
        \right]
        } \idh \\
        &= \int \calQ_{\calS} \log \frac{\calQ_{\calS} Z}{
                \prod\limits_{w \in \calS} \calN(; \hat{\bfo}_{w}, \sigma^2_\calP I)
        } \idh \\
        &= \int \sum\limits_{w \in \calS} \calQ_{\calS} \log \frac{\calQ_{\calS} Z}{
            \calN(; \hat{\bfo}_{w}, \sigma^2_\calP I)
        } \idh \\
        &= \int \sum\limits_{w \in \calS} \calQ_{\calS} \left( \log \calQ_{\calS} + \log Z -
            \log \calN(; \hat{\bfo}_{w}, \sigma^2_\calP I)
        \right) \idh \\
        &= \int \sum\limits_{w \in \calS} \calQ_{\calS} \left( \log \frac{\calQ_\calS}{\calN(; \hat{\bfo}_{w}, \sigma^2_\calP I)}
        + \log Z
        \right) \idh \\
        &= \sum\limits_{w \in \calS} \int \calQ_{\calS} \left( \log \frac{\calQ_\calS}{\calN(; \hat{\bfo}_{w}, \sigma^2_\calP I)}
        \right) \idh +
        \sum\limits_{w \in \calS} \int \calQ_{\calS} \log Z \idh \\
        &= \sum\limits_{w \in \calS} \int \calQ_{\calS} \left( \log \frac{\calQ_\calS}{\calN(; \hat{\bfo}_{w}, \sigma^2_\calP I)}
        \right) \idh +
        |\calS| \log Z\\
        &= \sum_{w \in \calS} \KL \left(
            \calQ_\calS || \calN (; \hat{\bfo}_w, \sigma^2_\calP I )
        \right)
        + |\calS| \log Z,
        \label{eq:explicit-kl-poe}
\end{align}
where $Z = \int \prod\limits_{w \in \calS} \calN(\bfh; \hat{\bfo}_{w}, \sigma^2_\calP I) \idh$.
The only first term in~\cref{eq:explicit-kl-poe} contributes the loss function because $|\calS| \log Z$ is constant when $\calS$ is fixed.
Therefore,
\begin{align}
    \sum_{w \in \calS} \KL \left(
        \calQ_\calS || \calN (; \hat{\bfo}_w, \sigma^2_\calP I )
    \right)
    = & \frac{1}{2 \sigma^2_\calP} \sum_{w \in \calS} \left(
        \left|\left| \boldsymbol{\mu}_{\calS} \right|\right| _2^2
        - 2 \boldsymbol{\mu}_{\calS}^{\top} \hat{\bfo}_w
        + \left|\left| \hat{\bfo}_w \right|\right| _2^2
    \right) \nonumber \\
    & + \frac{|\calS|}{2} \left\{\ln \frac{| \sigma^2_\calP I |}{ | \sigma^2_\calQ I |} - d  + \mathrm{Tr} \left[(\sigma_\calP^2 I)^{-1} \sigma_\calQ^2 I \right] \right\} \\
    =& \frac{1}{2\sigma^2_\calP}  \left(
        |\calS| \times \left|\left| \boldsymbol{\mu}_{\calS} \right|\right| _2^2
        - 2 \boldsymbol{\mu}_{\calS}^{\top} \sum_{w \in \calS}  \hat{\bfo}_w
        + \sum_{w \in \calS}  \left|\left| \hat{\bfo}_w \right|\right|_2^2
    \right) \nonumber \\
    & + \frac{|\calS|}{2} \left( d \ln \frac{\sigma^2_\calP}{\sigma^2_\calQ} - d  + d \frac{\sigma_\calQ^2}{\sigma_\calP^2} \right) \\
    =& \frac{|\calS|}{2\sigma^2_\calP} \left(
            \left|\left| \boldsymbol{\mu}_{\calS} \right|\right|_2^2
            - 2 \boldsymbol{\mu}_{\calS}^{\top} \frac{1}{|\calS|}\sum_{w \in \calS} \hat{\bfo}_w
        \right)
        + \frac{1}{2\sigma^2_\calP} \sum_{w \in \calS} \left|\left| \hat{\bfo}_w \right|\right| _2^2
        \nonumber \\
        & + \frac{|\calS| d}{2} \left( \ln \frac{\sigma^2_\calP}{\sigma^2_\calQ} - 1 + \frac{\sigma_\calQ^2}{\sigma_\calP^2} \right) \\
    =&
    \frac{|\calS|}{2\sigma^2_\calP} \left|\left|
        \boldsymbol{\mu}_{\calS} - \frac{1}{|\calS|}\sum_{w \in \calS} \hat{\bfo}_w
        \right|\right|_2^2
        + \frac{|\calS| d}{2} \left( \ln \frac{\sigma^2_\calP}{\sigma^2_\calQ} + \frac{\sigma_\calQ^2}{\sigma_\calP^2} \right) \nonumber \\
            &   - \frac{|\calS|d}{2}
                - \frac{1}{2\sigma^2_\calP} \left(
                |\calS| \left|\left| \frac{1}{|\calS|}\sum_{w \in \calS} \hat{\bfo}_w \right|\right|_2^2
                - \sum_{w \in \calS} \left|\left| \hat{\bfo}_w \right|\right|_2^2
                \right).
    \label{eq:kl_term_one_sentence}
\end{align}
Regarding minimization for the loss function by updating $\calQ$,
we can ignore the third and the firth terms in~\cref{eq:kl_term_one_sentence}, so we replace them with $C$.
\end{proof}

\subsection{Details of \cref{eq:avg} and \cref{eq:variance-avg} \label{sec:losses}}
Given bounded squared loss function $\ell(x, y, h) = \frac{1}{2} || y - \bfh_\calS ||$,
we minimize the upper bound based on~\cref{eq:general_loss} defined by
\begin{align}
    L(\calD_S, \calQ_\calS)
    =& \bbE_{h \sim \calQ_\calS} \frac{1}{N} \sum_{i=1}^{N}
        \left( \frac{1}{2} \left|\left| \hat{\bfi}_{w_i} - \mathbf{h}_\calS \right|\right|_2^2 \right)
        + \frac{|\calS|}{2 \sigma^2_\calP \lambda} \left|\left| \boldsymbol{\mu}_\calS - \frac{1}{|\calS|} \sum_{w \in \calS} \hat{\bfo}_w \right|\right|_2^2
        + \frac{|\calS| d}{2 \lambda} \left( \ln \frac{\sigma^2_\calP}{\sigma^2_\calQ} + \frac{\sigma_\calQ^2}{\sigma_\calP^2} \right) + C.
        \label{eq:squared_loss}
\end{align}
In our setting, $\calD_S = \{(\bfx_w, \hat{\bfh}_w) \mid w \in \calS \}$ in~\cref{eq:squared_loss},
and we use reparameterization trick for $\mathbf{h}_\calS$ in the first term,
\begin{align}
    L(D_\calS, \calQ_{\calS})
        =& \bbE_{h \sim \calQ_{\calS}}\frac{1}{|\calS|} \sum_{w \in \calS}
            \left( \frac{1}{2} \left|\left| \hat{\bfi}_{w} - \mathbf{h}_\calS \right|\right|_2^2 \right)
            +
            \frac{|\calS|}{2 \sigma^2_\calP \lambda} \left|\left| \boldsymbol{\mu}_\calS - \frac{1}{|\calS|} \sum_{w \in \calS} \hat{\bfo}_w \right|\right|_2^2
            + \frac{|\calS| d}{2 \lambda} \left( \ln \frac{\sigma^2_\calP}{\sigma^2_\calQ} + \frac{\sigma_\calQ^2}{\sigma_\calP^2} \right)  + C\\
            =& \bbE_{\boldsymbol{\epsilon} \sim \calN(\boldsymbol{\epsilon}; \mathbf{0}, I )}
            \frac{1}{|\calS|} \sum_{w \in \calS}
                \left( \frac{1}{2} \left|\left| \hat{\bfi}_{w} - \boldsymbol{\mu}_\calS - \sqrt{\sigma^2_\calQ} \boldsymbol{\epsilon} \right|\right|_2^2 \right) \nonumber \\
            &+ \frac{|\calS|}{2 \sigma^2_\calP \lambda} \left|\left| \boldsymbol{\mu}_\calS - \frac{1}{|\calS|} \sum_{w \in \calS} \hat{\bfo}_w \right|\right|_2^2
            + \frac{|\calS| d}{2 \lambda} \left( \ln \frac{\sigma^2_\calP}{\sigma^2_\calQ} + \frac{\sigma_\calQ^2}{\sigma_\calP^2} \right) + C.
            \label{eq:squared_loss_one_sentence}
\end{align}

We take derivative of \cref{eq:squared_loss_one_sentence} with respect to $\boldsymbol{\mu}_\calS$, and then set it to zero.
\begin{align}
    \nabla_{\boldsymbol{\mu}_\calS} L(\calD_S, \calQ_{\calS}) & = 0 \\
    \bbE_{\boldsymbol{\epsilon} \sim \calN(\boldsymbol{\epsilon}; \mathbf{0}, I )}
    \frac{1}{|\calS|} \sum_{w \in \calS}
    \left(\sqrt{\sigma^2_{\calQ}} \boldsymbol{\epsilon} + \boldsymbol{\mu}_\calS - \hat{\bfi}_{w}  \right)
        + \frac{|\calS|}{\sigma^2_\calP \lambda} \left( \boldsymbol{\mu}_\calS - \frac{1}{|\calS|} \sum_{w \in \calS} \hat{\bfo}_w  \right)
    & = 0 \\
    \sqrt{\sigma^2_{\calQ}} \bbE_{\boldsymbol{\epsilon} \sim \calN(\boldsymbol{\epsilon}; \mathbf{0}, I )} \boldsymbol{\epsilon} + \sum_{w \in \calS}
    \frac{1}{|\calS|}  \left(\boldsymbol{\mu}_\calS - \hat{\bfi}_{w}  \right)
        + \frac{|\calS|}{\sigma^2_\calP \lambda} \left( \boldsymbol{\mu}_\calS - \frac{1}{|\calS|} \sum_{w \in \calS} \hat{\bfo}_w  \right)
    & = 0 \\
    \boldsymbol{\mu}_\calS - \frac{1}{|\calS|} \sum_{w \in \calS} \hat{\bfi}_{w}
        + \alpha \boldsymbol{\mu}_\calS - \frac{\alpha}{|\calS|} \sum_{w \in \calS} \hat{\bfo}_w
    & = 0 \\
    (1 + \alpha) \boldsymbol{\mu}_\calS - \frac{1}{|\calS|} \sum_{w \in \calS}  \hat{\bfi}_{w}
        - \frac{\alpha}{|\calS|} \sum_{w \in \calS} \hat{\bfo}_w
    & = 0 \\
    (1 + \alpha) \boldsymbol{\mu}_\calS -
    \frac{1}{|\calS|} \sum_{w \in \calS} \left( \hat{\bfi}_{w} + \alpha \hat{\bfo}_w  \right)
    & = 0 \\
    (1 + \alpha) \boldsymbol{\mu}_\calS
    & = \frac{1}{|\calS|} \sum_{w \in \calS} \left( \hat{\bfi}_{w} + \alpha \hat{\bfo}_w \right) \\
    \boldsymbol{\mu}_\calS
    & = \frac{1}{(1 + \alpha) |\calS|}\sum_{w \in \calS} \left( \hat{\bfi}_{w} + \alpha \hat{\bfo}_w  \right),
\end{align}
where $\alpha = \frac{|\calS|}{\sigma^2_\calP \lambda}$.

We also take derivative of \cref{eq:squared_loss_one_sentence} with respect to $\sigma^2_\calQ$, and set it to zero.
\begin{align}
    \nabla_{\sigma^2_\calQ} L(\calD_S, \calQ_{\calS}) & = 0 \\
    \sqrt{\sigma^2_{\calQ}} \bbE_{\boldsymbol{\epsilon} \sim \calN(\boldsymbol{\epsilon}; \mathbf{0}, I )} \boldsymbol{\epsilon}
    + \frac{|\calS| d}{2 \lambda} \left(
        - \frac{1}{\sigma^2_\calQ} + \frac{1}{\sigma_\calP^2}
    \right)
    & = 0 \\
    - \frac{1}{\sigma^2_\calQ} + \frac{1}{\sigma_\calP^2}
    & = 0 \\
    \sigma^2_\calQ & = \sigma_\calP^2.
\end{align}

\subsection{Inverse Document Frequency Weighing from PAC-Bayes Bound \label{sec:idf}}
We start from~\cref{eq:explicit-kl-poe}.
Each variance $\sigma^2_{\calP_w}$ of priors depends on each word $w$.
\begin{align}
    \KL(\calQ_\calS || \calP_\calS)
    &=
    \sum_{w \in \calS} \KL \left(
        \calQ_\calS || \calN (; \hat{\bfo}_w, \sigma^2_{\calP_w} I)
    \right) \\
    &= \sum_{w \in \calS} \left[ \frac{1}{2 \sigma^2_{\calP_w}} \left(
        \left|\left| \bfh_{\calS} \right|\right|_2^2
        - 2 \bfh_{\calS}^{\top} \hat{\bfo}_w
        + \left|\left| \hat{\bfo}_w \right|\right|_2^2
    \right) +
    \frac{d}{2} \left( \ln \frac{\sigma^2_{\calP_w}}{\sigma^2_{\calQ}} - 1 + \frac{\sigma_\calQ^2}{\sigma_{\calP_w}^2} \right) \right] \\
    &=
        \left|\left| \bfh_{\calS} \right|\right|_2^2 \sum_{w \in \calS} \frac{1}{2 \sigma^2_{\calP_w}}
        - 2 \bfh_{\calS}^{\top} \sum_{w \in \calS} \frac{1}{2 \sigma^2_{\calP_w}} \hat{\bfo}_w
        + \sum_{w \in \calS}  \frac{1}{2 \sigma^2_{\calP_w}} \left|\left| \hat{\bfo}_w \right|\right|_2^2
        + \sum_{w \in \calS} \frac{d}{2} \left( \ln \frac{\sigma^2_{\calP_w}}{\sigma^2_{\calQ}} - 1 + \frac{\sigma_\calQ^2}{\sigma_{\calP_w}^2} \right)
    \\
    &= \eta \left|\left| \bfh_{\calS} \right|\right|_2^2
        - 2 \bfh_{\calS}^{\top} \sum_{w \in \calS} \frac{1}{2 \sigma^2_{\calP_w}} \hat{\bfo}_w
    + \sum_{w \in \calS} \frac{d}{2} \left( - \ln \sigma^2_{\calQ} + \frac{\sigma_\calQ^2}{\sigma_{\calP_w}^2} \right)
    - |\calS| \frac{d}{2}
    \\
    &= \eta \left(
            \left|\left| \bfh_{\calS} \right|\right|_2^2
            - \frac{2}{\eta} \bfh_{\calS}^{\top} \sum_{w \in \calS} \frac{1}{2 \sigma^2_{\calP_w}} \hat{\bfo}_w
        \right)
        + \sum_{w \in \calS} \frac{d}{2} \left( - \ln \sigma^2_{\calQ}  + \frac{\sigma_\calQ^2}{\sigma_{\calP_w}^2} \right)
        - |\calS| \frac{d}{2}
    \\
    &= \eta \left|\left|
        \bfh_{\calS} - \frac{1}{\eta} \sum_{w \in \calS} \frac{1}{2 \sigma^2_{\calP_w}} \hat{\bfo}_w
    \right|\right|^2_2
    + \sum_{w \in \calS} \frac{d}{2} \left( - \ln \sigma^2_{\calQ}  + \frac{\sigma_\calQ^2}{\sigma_{\calP_w}^2} \right)
    - \left(
        |\calS| \frac{d}{2}
        + \eta \left|\left|\frac{1}{\eta} \sum_{w \in \calS} \frac{1}{2 \sigma^2_{\calP_w}} \hat{\bfo}_w  \right|\right|_2^2
    \right),
\end{align}
where $\eta = \sum_{w \in \calS} \frac{1}{2 \sigma^2_{\calP_w}}$.
The last term can be ignored because it does not depend on posterior parameters $\boldsymbol{\mu}_\calS$ and $\sigma^2_{\calQ}$, so we replace it with $C$.

The weighted loss function is defined by $\ell(\bfx_i, y_i, h) = \frac{\beta(\bfx_i)}{2} || \hat{\bfi}_{w_i} - \bfh_{\calS} ||_2^2$,
where $\beta$ is a weighing function from $\bfx$ to $\bbR_+$.
We abbreviate $\beta(\mathbf{x}_w)$ to $\beta_w$.
We follow the same way as~\cref{sec:losses}.
\begin{align}
    L(\calD_S, \calQ_\calS)
        =& \bbE_{h \sim \calQ_\calS}
        \frac{1}{N} \sum_{i=1}^{N} \ell(\bfx_i, y_i, h)
            + \frac{1}{\lambda} \KL(\calQ_\calS || \calP_{\calS}) \\
            =& \bbE_{h \sim \calQ_\calS} \frac{1}{|\calS|} \sum_{w \in \calS}
            \left( \frac{\beta_{w}}{2} \left|\left| \hat{\bfi}_{w} - \mathbf{h}_{\calS}\right|\right|_2^2 \right)
            \nonumber \\
        & + \frac{1}{\lambda} \eta
            \left|\left|
                \boldsymbol{\mu}_\calS
                - \frac{1}{\eta} \sum_{w \in \calS} \frac{1}{2 \sigma^2_{\calP_w}} \hat{\bfo}_w
            \right|\right|^2_2
            + \sum_{w \in \calS} \frac{d }{2\lambda} \left( - \ln \sigma^2_{\calQ}  + \frac{\sigma_\calQ^2}{\sigma_{\calP_w}^2} \right)
            + C \\
        =& \bbE_{\boldsymbol{\epsilon} \sim \calN(\boldsymbol{\epsilon} ;\mathbf{0}, I)} \frac{1}{|\calS|} \sum_{w \in \calS}
            \left( \frac{\beta_{w}}{2} \left|\left| \hat{\bfi}_{w} - \boldsymbol{\mu}_{\calS} - \sqrt{\sigma^2_\calQ} \boldsymbol{\epsilon} \right|\right|_2^2 \right) \nonumber \\
        & + \frac{1}{\lambda} \eta
            \left|\left|
                \boldsymbol{\mu}_\calS
                - \frac{1}{\eta} \sum_{w \in \calS} \frac{1}{2 \sigma^2_{\calP_w}} \hat{\bfo}_w
            \right|\right|^2_2
            + \sum_{w \in \calS} \frac{d }{2\lambda} \left( - \ln \sigma^2_{\calQ} + \frac{\sigma_\calQ^2}{\sigma_{\calP_w}^2} \right)
    + C.
        \label{eq:idf-explict-loss}
\end{align}

We take derivative of~\cref{eq:idf-explict-loss} with respect to $\boldsymbol{\mu}_\calS$, and then set it to zero.
\begin{align}
    \nabla_{\boldsymbol{\mu}_{\calS}} L(\calD_S, \calQ_\calS) & = 0 \\
    \bbE_{\boldsymbol{\epsilon} \sim \calN(\boldsymbol{\epsilon}; \mathbf{0}, I)} \frac{1}{|\calS|} \sum_{w \in \calS} \beta_{w}
    \left( \sqrt{\sigma^2_\calQ} \boldsymbol{\epsilon} + \boldsymbol{\mu}_{\calS} - \hat{\bfi}_{w}  \right)
        + 2 \frac{1}{\lambda} \eta \left( \boldsymbol{\mu}_\calS
        - \frac{1}{\eta} \sum_{w \in \calS} \frac{1}{2 \sigma^2_{\calP_w}} \hat{\bfo}_w
        \right)
    & = 0 \\
    \frac{1}{|\calS|} \sum_{w \in \calS} \beta_w \boldsymbol{\mu}_\calS - \frac{1}{|\calS|} \sum_{w \in \calS} \beta_w \hat{\bfi}_{w}
        + 2 \frac{1}{\lambda} \eta \boldsymbol{\mu}_\calS
        - 2 \frac{1}{\lambda} \sum_{w \in \calS} \frac{1}{2 \sigma^2_{\calP_w}} \hat{\bfo}_w
    & = 0 \\
    \left(
        \frac{\sum_{w \in \calS} \beta_w }{|\calS|} + 2 \frac{1}{\lambda} \eta
    \right)
    \boldsymbol{\mu}_\calS - \frac{1}{|\calS|} \sum_{w \in \calS} \beta_w \hat{\bfi}_{w}
        - 2 \frac{1}{\lambda} \sum_{w \in \calS} \frac{1}{2 \sigma^2_{\calP_w}} \hat{\bfo}_w
    & = 0 \\
    \left(
        \sum_{w \in \calS} \beta_w + 2 \eta \frac{1}{\lambda} |\calS|
    \right)
    \boldsymbol{\mu}_\calS -
        \sum_{w \in \calS}
        \left( \beta_w \hat{\bfi}_{w}
            + \frac{1}{\lambda} |\calS| \frac{1}{\sigma^2_{\calP_w}} \hat{\bfo}_w
        \right)
    &= 0\\
    \left(
            \sum_{w \in \calS} \beta_w + 2 \eta \frac{1}{\lambda}|\calS|
    \right)
    \boldsymbol{\mu}_\calS & =
        \sum_{w \in \calS}
            \left( \beta_w \hat{\bfi}_{w}
                + \frac{1}{\lambda} |\calS| \frac{1}{\sigma^2_{\calP_w}} \hat{\bfo}_w
            \right) \\
    \boldsymbol{\mu}_\calS & =
    \frac{
        \sum_{w \in \calS}
            \beta_w \hat{\bfi}_{w}
                + \frac{1}{\lambda} |\calS| \frac{1}{\sigma^2_{\calP_w}} \hat{\bfo}_w
        }{
            \sum_{w \in \calS} \beta_w + 2 \eta \frac{1}{\lambda} |\calS|
        }.
\end{align}
We set $\frac{1}{\sigma^2_{\calP_w}} = \mathrm{IDF}(w)$.
Then, we evaluate $\eta = \frac{1}{2} \sum_{w \in \calS} \frac{1}{\sigma^2_{\calP_w}} = \frac{1}{2} \sum_{w \in \calS} \mathrm{IDF}(w)$, then
\begin{align}
    \boldsymbol{\mu}_{\calS}
    & = \frac{
            \sum_{w \in \calS}
                \beta_w \hat{\bfi}_{w} + \frac{1}{\lambda} |\calS| \mathrm{IDF}(w) \hat{\bfo}_w
        }{
            \sum_{w \in \calS}
                \beta_w + \frac{1}{\lambda} |\calS| \sum_{w \in \calS} \mathrm{IDF}(w)
        }
    \\
    & = \frac{
        \sum_{w \in \calS} \beta_w \hat{\bfi}_{w} + \frac{1}{\lambda} |\calS| \mathrm{IDF}(w) \hat{\bfo}_w
    }{
        \sum_{w \in \calS}
            \beta_w + \frac{1}{\lambda} |\calS| \mathrm{IDF}(w)
    }.
\end{align}
We set $\beta_w = |\calS| \mathrm{IDF}(w)$,
\begin{align}
    \boldsymbol{\mu}_{\calS}
    & = \frac{
        \sum_{w \in \calS}
            |\calS| \mathrm{IDF}(w) \hat{\bfi}_{w} + \frac{1}{\lambda} |\calS| \mathrm{IDF}(w) \hat{\bfo}_w
    }{
        \sum_{w \in \calS}
            |\calS| \mathrm{IDF}(w) + \frac{1}{\lambda} |\calS|  \mathrm{IDF}(w)
    }
    \\
    & = \frac{
        \sum_{w \in \calS} \mathrm{IDF}(w) \hat{\bfi}_{w} + \frac{1}{\lambda}\mathrm{IDF}(w) \hat{\bfo}_w
    }{
        \sum_{w \in \calS}
            \mathrm{IDF}(w) + \frac{1}{\lambda} \mathrm{IDF}(w)
    }
    \\
    & = \frac{
        \sum_{w \in \calS} \mathrm{IDF}(w) \left(
            \hat{\bfi}_{w} + \frac{1}{\lambda} \hat{\bfo}_w
        \right)
    }{
        \left(1 + \frac{1}{\lambda} \right) \sum_{w \in \calS} \mathrm{IDF}(w)
    }.
\end{align}

We also take derivative of~\cref{eq:idf-explict-loss} with respect to $\sigma_\calQ^2$, and then set it to zero.
\begin{align}
    \nabla_{\sigma^2_\calQ} L(\calD_S, \calQ_{\calS}) & = 0 \\
    \sqrt{\sigma^2_{\calQ}} \bbE_{\boldsymbol{\epsilon} \sim \calN(;\mathbf{0}, I )} \boldsymbol{\epsilon}
    + \sum_{w \in \calS} \frac{d }{2\lambda} \left( - \frac{1}{\sigma^2_{\calQ}}  + \frac{1}{\sigma_{\calP_w}^2} \right)
    & = 0 \\
    - \sum_{w \in \calS} \left( \frac{1}{\sigma^2_\calQ} - \frac{1}{\sigma_{\calP_w}^2} \right)
    & = 0 \\
    \sigma^2_\calQ & = \left( \frac{1}{|\calS|} \sum_{w \in \calS} \frac{1}{\sigma^2_{\calP_w}} \right)^{-1}.
\end{align}

\end{document}